\documentclass[11pt,a4paper]{article}
\usepackage[hyperref]{acl2020}
\usepackage{CJKutf8}
\usepackage{times}
\usepackage{hyperref}
\usepackage{amsmath}
\usepackage{graphicx}
\usepackage{latexsym}
\usepackage{amsfonts}
\usepackage{multirow}

\usepackage{helvet}
\usepackage{courier}
\usepackage{dashrule}
\usepackage{subfigure}
\usepackage{makecell}

\usepackage{microtype}

\aclfinalcopy 
\def\aclpaperid{797} 

\title{Spelling Error Correction with Soft-Masked BERT}

\author{Shaohua Zhang\textsuperscript{1}, Haoran Huang\textsuperscript{1}, Jicong Liu\textsuperscript{2} and Hang Li\textsuperscript{1}\\
 \textsuperscript{1}ByteDance AI Lab \\
 \textsuperscript{2}School of Computer Science and Technology, Fudan University \\
 \texttt{\{zhangshaohua.cs,huanghaoran,lihang.lh\}@bytedance.com} \\
  \texttt{jcliu15@fudan.edu.cn} \\}

\date{}
\begin{document}

\maketitle

\begin{abstract}
Spelling error correction is an important yet challenging task because a satisfactory solution of it essentially needs human-level language understanding ability. Without loss of generality we consider Chinese spelling error correction (CSC) in this paper. A state-of-the-art method for the task selects a character from a list of candidates for correction (including non-correction) at each position of the sentence on the basis of BERT, the language representation model. The accuracy of the method can be sub-optimal, however, because BERT does not have sufficient capability to detect whether there is an error at each position, apparently due to the way of pre-training it using mask language modeling. In this work, we propose a novel neural architecture to address the aforementioned issue, which consists of a network for error detection and a network for error correction based on BERT, with the former being connected to the latter with what we call soft-masking technique. Our method of using `Soft-Masked BERT' is general, and it may be employed in other language detection-correction problems. Experimental results on two datasets demonstrate that the performance of our proposed method is significantly better than the baselines including the one solely based on BERT.

\end{abstract}

\section{Introduction}
Spelling error correction is an important task which aims to correct spelling errors in a text either at word-level or at character-level~\cite{yu-li-2014-chinese,yu2014overview,zhang2015hanspeller++,wang2018hybrid,hong2019faspell,wang2019confusionset}. It is crucial for many natural language applications such as  search~\cite{DBLP:conf/tal/MartinsS04,DBLP:conf/coling/GaoLMQS10}, optical character recognition (OCR)~\cite{afli2016using,wang2018hybrid}, and essay scoring~\cite{burstein1999automated}. In this paper, we consider Chinese spelling error correction (CSC) at character-level.

\begin{table}[t]
\caption{Examples of Chinese spelling errors}
{
\hrule

Wrong: \begin{CJK*}{UTF8}{gbsn}埃及有金{\color{red}子}塔。\end{CJK*}Egypt has golden towers.\\
\hdashrule{8cm}{1pt}{3pt}\\
Correct: \begin{CJK*}{UTF8}{gbsn}埃及有金{\color{red}字}塔。\end{CJK*}Egypt has pyramids. \\
\hrule
Wrong: \begin{CJK*}{UTF8}{gbsn}他的求{\color{red}胜}欲很强，为了越狱在挖洞。\end{CJK*}
He has a strong desire to win and is digging for prison breaks\\
\hdashrule{8cm}{1pt}{3pt}\\
Correct: \begin{CJK*}{UTF8}{gbsn}他的求{\color{red}生}欲很强，为了越狱在挖洞。\end{CJK*}
He has a strong desire to survive and is digging for prison breaks.
\hrule
}
\end{table}\label{tb:examples}

Spelling error correction is also a very challenging task, because to completely solve the problem the system needs to have human-level language understanding ability. 
There are at least two challenges here, as shown in Table \ref{tb:examples}. First, world knowledge is needed for spelling error correction. Character \begin{CJK*}{UTF8}{gbsn}字\end{CJK*} in the first sentence is mistakenly written as \begin{CJK*}{UTF8}{gbsn}子\end{CJK*}, where \begin{CJK*}{UTF8}{gbsn}金子塔\end{CJK*} means golden tower and \begin{CJK*}{UTF8}{gbsn}金字塔\end{CJK*} means pyramid. Humans can correct the typo by referring to world knowledge. 
Second, sometimes inference is also required. In the second sentence, the 4-th character \begin{CJK*}{UTF8}{gbsn}生\end{CJK*} is mistakenly written as \begin{CJK*}{UTF8}{gbsn}胜\end{CJK*}. In fact, \begin{CJK*}{UTF8}{gbsn}胜\end{CJK*} and the surrounding characters form a new valid word \begin{CJK*}{UTF8}{gbsn}求胜欲\end{CJK*} (desire to win), rather than the intended word \begin{CJK*}{UTF8}{gbsn}求生欲\end{CJK*} (desire to survive). 

Many methods have been proposed for CSC or more generally spelling error correction. Previous approaches can be mainly divided into two categories. One employs traditional machine learning and the other deep learning~\cite{yu2014overview,tseng2015introduction,wang2018hybrid}. Zhang et al.~\shortcite{zhang2015hanspeller++}, for example, proposed a unified framework for CSC consisting of a pipeline of error detection, candidate generation, and final candidate selection using traditional machine learning. Wang et al.~\shortcite{wang2019confusionset} proposed a Seq2Seq model with copy mechanism which transforms an input sentence into a new sentence with spelling errors corrected.

More recently,  BERT~\cite{devlin2018bert}, the language representation model, is successfully applied to many language understanding tasks including CSC (cf.,~\cite{hong2019faspell}). In the state-of-the-art method using BERT, a character-level BERT is first pre-trained using a large unlabelled dataset and then fine-tuned using a labeled dataset. The labeled data can be obtained via data augmentation in which examples of spelling errors are generated using a large confusion table. Finally the model is utilized to predict the most likely character from a list of candidates at each position of the given sentence. The method is powerful because BERT has certain ability to acquire knowledge for language understanding. Our experimental results show that the accuracy of the method can be further improved, however. One observation is that the error detection capability of the model is not sufficiently high, and once an error is detected, the model has a better chance to make a right correction. We hypothesize that this might be due to the way of pre-training BERT with mask language modeling in which only about 15\% of the characters in the text are masked, and thus it only learns the distribution of masked tokens and tends to choose not to make any correction. This phenomenon is prevalent and represents a fundamental challenge for using BERT in certain tasks like spelling error correction.

To address the above issue, we propose a novel neural architecture in this work, referred to as Soft-Masked BERT. Soft-Masked BERT contains two networks, a detection network and a correction network based on BERT. The correction network is similar to that in the method of solely using BERT. The detection network is a Bi-GRU network that 
predicts the probability that the character is an error at each position. The probability is then utilized to conduct soft-masking of embedding of character at the position. Soft masking is an extension of conventional `hard masking' in the sense that the former degenerates to the latter, when the probability of error equals one. The soft-masked embedding at each position is then inputted into the correction network. The correction network conducts error correction using BERT. This approach can force the model to learn the right context for error correction under the help of the detection network, during end-to-end joint training.

We conducted experiments to compare Soft-Masked BERT and several baselines including the method of using BERT alone. As datasets we utilized the benchmark dataset of SIGHAN.  We also created a large and high-quality dataset for evaluation named News Title. The dataset, which contains titles of news articles, is ten times larger than the previous datasets. Experimental results show that Soft-Masked BERT significantly outperforms the baselines on the two datasets in terms of accuracy measures.

The contributions of this work include (1) proposal of the novel neural architecture Soft-Masked BERT for the CSC problem, (2) empirical verification of the effectiveness of Soft-Masked BERT. 

\section{Our Approach}
\begin{figure*}
  \centering
  \includegraphics[width=\textwidth]{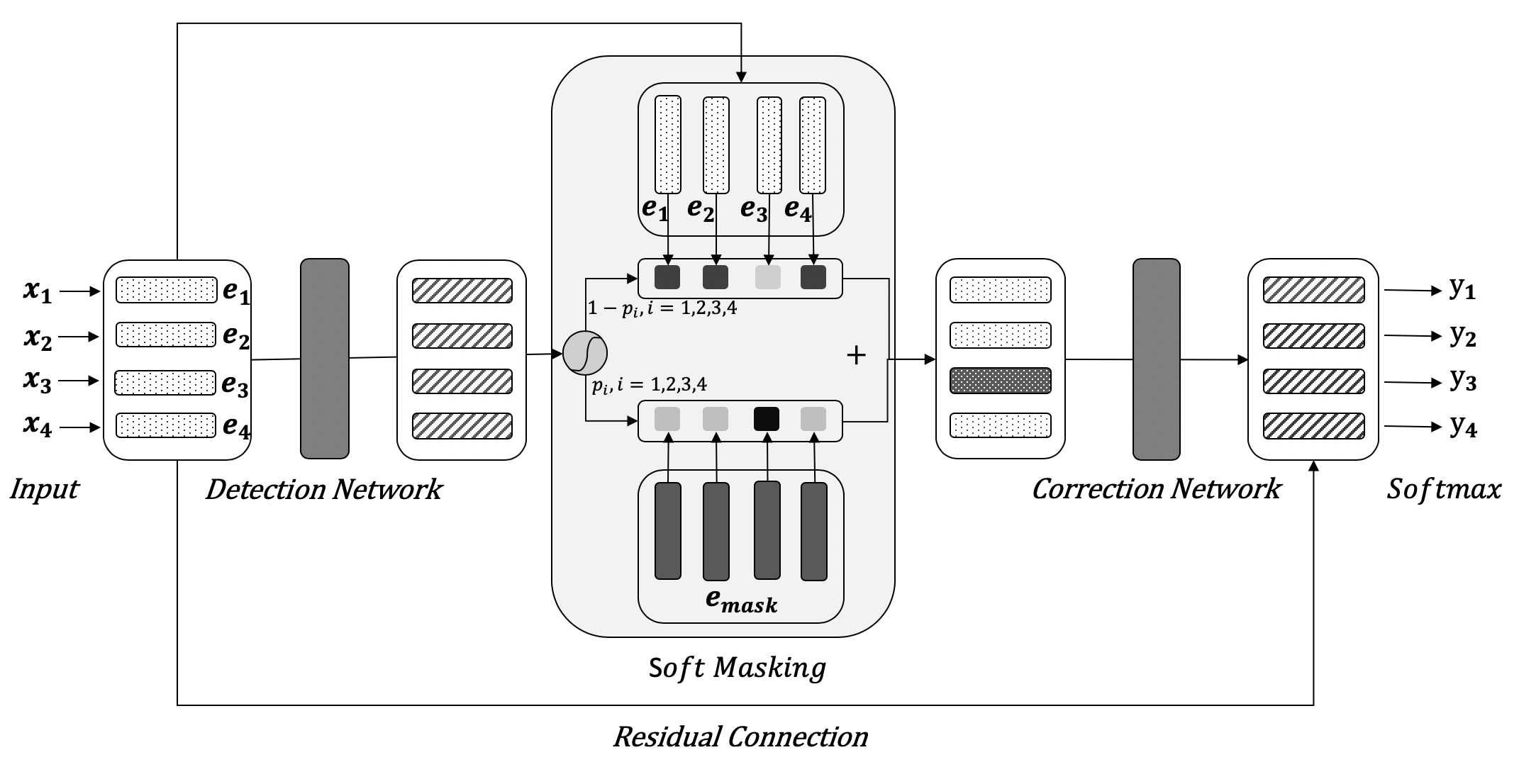}
  \caption{Architecture of Soft-Masked BERT} 
  \label{network} 
\end{figure*}

\subsection{Problem and Motivation}

Chinese spelling error correction (CSC) can be formalized as the following task. Given a sequence of $n$ characters (or words) $X=(x_1,x_2,\cdots,x_n)$, the goal is to transform it into another sequence of characters $Y=(y_1,y_2,\cdots,y_n)$ with the same length, where the incorrect characters in $X$ are replaced with the correct characters to obtain $Y$. The task can be viewed as a sequential labeling problem in which the model is a mapping function $f:X \rightarrow Y$. The task is an easier one, however, in the sense that usually no or only a few characters need to be replaced and all or most of the characters should be copied.

The state-of-the-art method for CSC is to employ BERT to accomplish the task.  Our preliminary experiments show that the performance of the approach can be improved, if the erroneous characters are designated (cf., section 3.6). In general the BERT based method tends to make no correction (or just copy the original characters).
Our interpretation is that in pre-training of BERT only 15\% of the characters are masked for prediction, resulting in learning of a model which does not possess enough capacity for error detection. This motives us to devise a new model.

\subsection{Model}

We propose a novel neural network model called Soft-Masked BERT for CSC, as illustrated in Figure \ref{network}. Soft-Masked BERT is composed of a detection network based on Bi-GRU and a correction network based on BERT. The detection network predicts the probabilities of errors and the correction network predicts the probabilities of error corrections, while the former passes its prediction results to the latter using soft masking.

More specifically, our method first creates an embedding for each character in the input sentence, referred to as input embedding. Next, it takes the sequence of embeddings as input and outputs the probabilities of errors for the sequence of characters (embeddings) using the detection network. After that it calculates the weighted sum of the input embeddings and [MASK] embeddings weighted by the error probabilities. The calculated embeddings {\em mask the likely errors in the sequence in a soft way}. Then, our method takes the sequence of soft-masked embeddings as input and outputs the probabilities of error corrections using the correction network, which is a BERT model whose final layer consists of a softmax function for all characters. There is also a residual connection between the input embeddings and the embeddings at the final layer. Next, we describe the details of the model. 

\subsection{Detection Network}

The detection network is a sequential binary labeling model. The input is the sequence of embeddings $E=(e_1,e_2,\cdots,e_n)$, where $e_i$ denotes the embedding of character $x_i$, which is the sum of word embedding, position embedding, and segment embedding of the character, as in BERT.
The output is a sequence of labels $ G = (g_1, g_2, \cdots, g_n) $, where $g_i$ denotes the label of the $i$ character, and 1 means the character is incorrect and 0 means it is correct. For each character there is a probability $p_i$ representing the likelihood of being 1.
The higher $p_i$ is the more likely the character is incorrect. 
 
In this work, we realize the detection network as a bidirectional GRU (Bi-GRU). For each character of the sequence, the probability of error $p_i$ is defined as
\begin{align}
   p_i =  P_d(g_i = 1 | X) = \sigma(W_{d}h_{i}^{d} + b_{d})
\end{align}
where $P_d(g_i = 1 | X)$ denotes the conditional probability given by the detection network, $\sigma$ denotes the sigmoid function, $h_{i}^{d}$ denotes the hidden state of Bi-GRU, $W_d$ and $b_d$ are parameters. Furthermore, the hidden state is defined as
\begin{align}
\overrightarrow{h_{i}^{d}} &= \text{GRU}(\overrightarrow{h}_{i-1}^{d}, e_{i})\\
\overleftarrow{h_{i}^{d}} &= \text{GRU}(\overleftarrow{h}_{i+1}^{d}, e_{i})\\
h_{i}^{d} &= [\overrightarrow{h_{i}^{d}};\overleftarrow{h_{i}^{d}}] 
\end{align}
where $[\overrightarrow{h_{i}^{d}};\overleftarrow{h_{i}^{d}}]$ denotes concatenation of GRU hidden states from the two directions and GRU is the GRU function.

Soft masking amounts to a weighted sum of input embeddings and mask embeddings with error probabilities as weights. The soft-masked embedding $e_{i}^{'}$ for the $i$-th character is defined as
\begin{align}
e_{i}^{'} = p_{i} \cdot e_{mask} + (1 - p_{i}) \cdot e_i 
\end{align}
where $e_i$ is the input embedding and $e_{mask}$ is the mask embedding. If the probability of error is high, then soft-masked embedding $e_{i}^{'}$ is close to the mask embedding $e_{mask}$; otherwise it is close to the input embedding $e_i$.

\subsection{Correction Network}

The correction network is a sequential multi-class labeling model based on BERT. The input is the sequence of soft-masked embeddings $E'=(e'_1,e'_2,\cdots,e'_n)$ and the output is a sequence of characters $ Y = (y_1, y_2, \cdots, y_n) $. 

BERT consists of a stack of 12 identical blocks taking the entire sequence as input. Each block contains a multi-head self-attention operation followed by a feed-forward network, defined as:
\begin{align}
\begin{split}
&\text{MultiHead}(Q,K,V)\\ 
&= \text{Concat}(\text{head}_1;
\cdots, \text{head}_h)W^{O}
\end{split}\\
&\text{head}_i = \text{Attention}(QW^{Q}_i, KW^{K}_i, VW_i^{V})\\
&\text{FFN}(X) = \text{max}(0, X W_1 + b_1)W_2 + b_2
\end{align}
where $Q$, $K$, and $V$ are the same matrices representing the input sequence or the output of the previous block, MultiHead, Attention, and FNN denote multi-head self-attention, self-attention, and feed-forward network respectively, $W^O$, $W_i^Q$, $W_i^K$, $W_i^V$, $W_1$, $W_2$, $b_1$, and $b_2$ are parameters. We denote the sequence of hidden states at the final layer of BERT as $H^{c}=(h_1^{c}, h_2^{c}, \cdots, h_n^{c})$

For each character of the sequence, the probability of error correction is defined as
\begin{align}
P_c(y_i = j|X) &= \text{softmax}(Wh_i^{'} + b)[j]
\end{align}
where $P_c(y_i = j|X)$ is the conditional probability that character $x_i$ is corrected as character $j$ in the candidate list, softmax is the softmax function, $h_{i}^{'}$ denotes the hidden state, $W$ and $b$ are parameters.
Here the hidden state $h_{i}^{'}$ is obtained by linear combination with the residual connection,
\begin{align}
h_{i}^{'} &= h_{i}^{c} + e_{i}
\end{align}
where $h_i^c$ is the hidden state at the final layer and $e_i$ is the input embedding of character $x_i$. 
The last layer of correction network exploits a softmax function. The character that has the largest probability is selected from the list of candidates as output for character $x_i$.

\subsection{Learning}

The learning of Soft-Masked BERT is conducted end-to-end, provided that BERT is pre-trained and training data is given which consists of pairs of original sequence and corrected sequence, denoted as $\mathcal{} = \{(X_1, Y_1),(X_2, Y_2),\ldots,(X_N, Y_N)\}$. One way to create the training data is to repeatedly generate a sequence $X_i$ containing errors given a sequence $Y_i$ without an error, using a confusion table, where $i=1,2,\cdots, N$.

The learning process is driven by optimizing two objectives, corresponding to error detection and error correction respectively.
\begin{align}
\mathcal{L}_d &= -\sum_{i=1}^{n}\log P_d(g_i|X)\\
\mathcal{L}_c &= -\sum_{i=1}^{n}\log P_c(y_i|X)
\end{align}
where $\mathcal{L}_d$ is the objective for training of the detection network, and $\mathcal{L}_c$ is the objective for training of the correction network (and also the final decision). The two functions are linearly combined as the overall objective in learning.
\begin{equation}
\mathcal{L} = \lambda \cdot \mathcal{L}_c + (1 - \lambda) \cdot \mathcal{L}_d
\end{equation}
where $\lambda \in [0,1]$ is coefficient.

\section{Experimental Results}
\subsection{Datasets}
We made use of the SIGHAN dataset, a benchmark for CSC\footnote{Following the common practice~\cite{wang2019confusionset}, we converted the characters in the dataset from traditional Chinese to simplified Chinese.}. 
SIGHAN is a small dataset containing 1,100 texts and 461 types of errors (characters). The texts are collected from the essay section of Test of Chinese as Foreign Language and the topics are in a narrow scope. We adopted the standard split of training, development, and test data of SIGHAN.

We also created a much larger dataset for testing and development, referred to as News Title. We sampled the titles of news articles at Toutiao, a Chinese news app with a large variety of content in politics, entertainment, sports, education, etc. To ensure that the dataset contains a sufficient number of incorrect sentences, we conducted the sampling from lower quality texts, and thus the error rate of the dataset is higher than usual. Three people conducted five rounds of labeling to carefully correct spelling errors in the titles. The dataset contains 15,730 texts. There are 5,423 texts containing errors, in 3,441 types. We divided the data into test set and development set, each containing 7,865 texts.

In addition, we followed the common practice in CSC to automatically generate a dataset for training. We first crawled about 5 million news titles at the Chinese news app. We also created a confusion table in which each character is associated with a number of homophonous characters as potential errors. Next, we randomly replaced 15\% of the characters in the texts with other characters to artificially generate errors, where 80\% of them are homophonous characters in the table and 20\% of them are random characters. This is because in practice about 80\% of spelling errors in Chinese are homophonous characters due to the use of Pinyin-based input methods by people.

\begin{table*}[t]
  \centering
  \caption{Performances of Different Methods on CSC}
  \begin{tabular}{|p{1.7cm}|c|c|c|c|c|c|c|c|c|}
    \hline
    \multirow{2}{*}{\textbf{Test Set}} & \multirow{2}{*}{\textbf{Method}} & \multicolumn{4}{|c|}{\textbf{Detection}} & \multicolumn{4}{|c|}{\textbf{Correction}} \\
    \cline{3-10}
    ~ & ~ & \textbf{Acc.} & \textbf{Prec.} & \textbf{Rec.} & \textbf{F1.} & \textbf{Acc.} & \textbf{Prec.} & \textbf{Rec.} & \textbf{F1.} \\
    \hline
    \hline
  \multirow{9}{*}{SIGHAN} & NTOU~\shortcite{tseng2015introduction} & 42.2 & 42.2 & 41.8 & 42.0 & 39.0 & 38.1 & 35.2 & 36.6\\
  \cline{2-10}
  ~ & NCTU-NTUT~\shortcite{tseng2015introduction}  & 60.1 & 71.7 & 33.6 & 45.7 & 56.4 & 66.3 & 26.1 & 37.5\\
  \cline{2-10}
  ~ & HanSpeller++~\shortcite{zhang2015hanspeller++}  & 70.1 & \textbf{80.3} & 53.3 & 64.0 & 69.2& \textbf{79.7} & 51.5& 62.5 \\
  \cline{2-10}
  ~ & Hybird~\shortcite{wang2018hybrid}  &- & 56.6 & 69.4& 62.3& -& - & -& 57.1 \\
  \cline{2-10}
  ~ & FASPell~\shortcite{hong2019faspell}  &74.2 & 67.6 & 60.0& 63.5& 73.7& 66.6 & 59.1& 62.6 \\
  \cline{2-10}
  ~ & Confusionset~\shortcite{wang2019confusionset}  & - & 66.8 & 73.1 & 69.8 & - & 71.5 & 59.5 & 64.9 \\
  \cline{2-10}
  ~ & BERT-Pretrain & 6.8 &3.6 & 7.0 & 4.7 & 5.2 & 2.0 & 3.8 & 2.6\\
  \cline{2-10}
  ~ & BERT-Finetune  &80.0&73.0&70.8&71.9&76.6&  65.9& 64.0& 64.9\\
  \cline{2-10}
  \cline{2-10}
  ~ & Soft-Masked BERT& \textbf{80.9} & 73.7 & \textbf{73.2} & \textbf{73.5} &\textbf{77.4} & 66.7 & \textbf{66.2} & \textbf{66.4}\\
  \hline
  \hline
  \multirow{3}{*}{News Title} & BERT-Pretrain & 7.1 & 1.3 & 3.6 & 1.9 & 0.6 & 0.6 & 1.6 & 0.8\\
  \cline{2-10}
  ~ & BERT-Finetune  &80.0 & 65.0 & 61.5 & 63.2 & 76.8 &55.3&52.3 & 53.8\\
  \cline{2-10}
  ~ & Soft-Masked BERT & \textbf{80.8} & \textbf{65.5} & \textbf{64.0} & \textbf{64.8} & \textbf{77.6} &  \textbf{55.8} & \textbf{54.5} & \textbf{55.2}\\
  \hline
  \end{tabular}
    
    \centering
    \label{tab:performance}
\end{table*}

\subsection{Baselines}

For comparison, we adopted the following methods as baselines. We report the results of the methods from their original papers.

\textbf{NTOU} is a method of using an n-gram model and a rule-based classifier~\cite{tseng2015introduction}. 
\textbf{NCTU-NTUT} is a method of utilizing word vectors and conditional random field~\cite{tseng2015introduction}. 
\textbf{HanSpeller++} is an unified framework employing a hidden Markov model to generate candidates and a filter to re-rank candidates~\cite{zhang2015hanspeller++}.
\textbf{Hybrid} uses a BiLSTM-based model trained on a generated dataset~\cite{wang2018hybrid}.
\textbf{Confusionset} is a Seq2Seq model consisting of a pointer network and copy mechanism~\cite{wang2019confusionset}.
\textbf{FASPell} adopts a Seq2Seq model for CSC employing BERT as a denoising auto-encoder and a decoder~\cite{hong2019faspell}.
\textbf{BERT-Pretrain} is the method of using a pre-trained BERT.
\textbf{BERT-Finetune} is the method of using a fine-tuned BERT.

\subsection{Experiment Setting}

As evaluation measures, we utilized sentence-level accuracy, precision, recall, and F1 score as in most of the previous work. 
We evaluated the accuracy of a method in both detection and correction. Obviously correction is more difficult than detection, because the former is dependent on the latter.

The pre-trained BERT model utilized in the experiments is the one provided at
\href{https://github.com/huggingface/transformers}{https://github.com/huggingface/transformers}.
In fine-tuning of BERT, we kept the default hyper-parameters and only fine-tuned the parameters using Adam.
In order to reduce the impact of training tricks, we did not use the dynamic learning rate strategy and maintained a learning rate $2e^{-5}$ in fine-tuning. The size of hidden unit in Bi-GRU is 256 and all models use a batch size of 320.

In the experiments on SIGHAN, for all BERT-based models, we first fine-tuned the model with the 5 million training examples and then continued the fine-tuning with the training examples in SIGHAN. 
We removed the unchanged texts in the training data to improve the efficiency. 
In the experiments on News Title, the models were fine-tuned only with the 5 million training examples.

The development sets were utilized for hyper-parameter tuning for both SIGHAN and News Title. The best value for hyper-parameter $\lambda$ was chosen for each dataset.

\begin{table*}[t]
  \centering
  \caption{Impact of Different Sizes of Training Data}
  \begin{tabular}{|c|c|c|c|c|c|c|c|c|c|}
    \hline
    \multirow{2}{*}{\textbf{Train Set}} & \multirow{2}{*}{\textbf{Method}} & \multicolumn{4}{|c|}{\textbf{Detection}} & \multicolumn{4}{|c|}{\textbf{Correction}} \\
    \cline{3-10}
    ~ & ~ & \textbf{Acc.} & \textbf{Prec.} & \textbf{Rec.} & \textbf{F1.} & \textbf{Acc.} & \textbf{Prec.} & \textbf{Rec.} & \textbf{F1.} \\
    \hline
    \multirow{2}{*}{500,000} & BERT-Finetune & 71.8 & 49.6 & 48.2 & 48.9 & 67.4 & 36.5 & 35.5 & 36.0\\
    \cline{2-10}
    ~& Soft-Masked BERT & \textbf{72.3} & \textbf{50.3} & \textbf{49.6} & \textbf{50.0} & \textbf{68.2} & \textbf{37.9} & \textbf{37.4} & \textbf{37.6}\\
    \hline
    \multirow{2}{*}{1,000,000} & BERT-Finetune & 74.2 & 54.7 & 51.3 & 52.9 & 70.0 & 41.6 & 39.0 & 40.3\\
    \cline{2-10}
    ~& Soft-Masked BERT & \textbf{75.3} & \textbf{56.3} & \textbf{54.2} & \textbf{55.2} & \textbf{71.1} & \textbf{43.6} & \textbf{41.9} & \textbf{42.7}\\
    \hline
    \multirow{2}{*}{2,000,000} & BERT-Finetune & 77.0 & 59.7 & 57.0 & 58.3 & 73.1 & 48.0 & 45.8 & 46.9\\
    \cline{2-10}
    ~& Soft-Masked BERT & \textbf{77.6} & \textbf{60.0} & \textbf{58.5} & \textbf{59.2} & \textbf{73.7} & \textbf{48.4} & \textbf{47.3} & \textbf{47.8}\\
    \hline
    \multirow{2}{*}{5,000,000} & BERT-Finetune & 80.0 & 65.0 & 61.5 & 63.2 & 76.8 & 55.3 & 52.3 & 53.8\\
    \cline{2-10}
    ~ & Soft-Masked BERT & \textbf{80.8} & \textbf{65.5} & \textbf{64.0} & \textbf{64.8} & \textbf{77.6} &  \textbf{55.8} & \textbf{54.5} & \textbf{55.2}\\

  \hline
  \end{tabular}
  
    \label{tab:size}
\end{table*}

\subsection{Main Results}

Table \ref{tab:performance} presents the experimental results of all methods on the two test datasets. From the table, one can observe that the proposed model Soft-Masked BERT significantly outperforms the baseline methods on both datasets. Particularly, on News Title, Soft-Masked BERT performs much better than the baselines in terms of all measures. The best results for recall of correction level on the News Title dataset are greater than 54\%, which means more than 54\% errors will be found and correction level precision are better than 55\%.

HanSpeller++ achieves the highest precision on SIGHAN, apparently because it can eliminate false detections with its large number of manually-crafted rules and features. 
Although the use of rules and features is effective, the method has high cost in development and may also have difficulties in generalization and adaptation. 
In some sense, it is not directly comparable with the other learning-based methods including Soft-Masked BERT. The results of all methods except Confusionset are at sentence level not at character level. (The results at character level can look better.) Nonetheless, Soft-Mask BERT still performs significantly better. 

The three methods of using BERT, Soft-Masked BERT, BERT-Finetune, and FASPell, perform better than the other baselines, while the method of BERT-Pretrain performs fairly poorly. The results indicate that BERT without fine-tuning (i.e., BERT-Pretrain) would not work and BERT with fine-tuning (i.e., BERT-Finetune, etc) can boost the performances remarkably. Here we see another successful application of BERT, which can acquire certain amount of knowledge for language understanding. Furthermore, Soft-Masked BERT can beat BERT-Finetune by large margins on both datasets. The results suggest that error detection is important for the utilization of BERT in CSC and soft masking is really an effective means.

\begin{table*}[t]
  \centering
  \caption{Ablation Study of Soft-Masked BERT on News Title}
  \begin{tabular}{|c|c|c|c|c|c|c|c|c|}
    \hline
    \multirow{2}{*}{\textbf{Method}} & \multicolumn{4}{|c|}{\textbf{Detection}} & \multicolumn{4}{|c|}{\textbf{Correction}} \\
    \cline{2-9}
     ~ & \textbf{Acc.} & \textbf{Prec.} & \textbf{Rec.} & \textbf{F1.} & \textbf{Acc.} & \textbf{Prec.} & \textbf{Rec.} & \textbf{F1.} \\
     \hline
     \hline
  \makecell[c]{BERT-Finetune\\+Force(Upper Bound)}& 89.9 & 75.6 & 90.3 & 82.3 & 82.9 & 58.4  & 69.8 & 63.6\\ 
  \hline
  Soft-Masked BERT & 80.8 & 65.5 & 64.0 & 64.8 &77.6 &  55.8 & 54.5 & 55.2\\
  \hline
  Soft-Masked BERT-R & 81.0 & 75.2 & 53.9 & 62.8 & 78.4 & 64.6 & 46.3 & 53.9\\
  \hline
  Rand-Masked BERT & 70.9 & 46.6 & 48.5 & 47.5 & 68.1 & 38.8 & 40.3 & 39.5\\
  \hline
  BERT-Finetune  &80.0 & 65.0 & 61.5 & 63.2 & 76.8 &55.3&52.3 & 53.8\\ 
  \hline
  Hard-Masked BERT (0.95)  & 80.6 & 65.3 & 63.2 & 64.2 & 76.7 & 53.6 & 51.8 & 52.7\\
  \hline
  Hard-Masked BERT (0.9)  & 77.4 & 57.8 & 60.3 & 59.0 & 72.4 & 44.0 & 45.8 & 44.9\\
  \hline
  Hard-Masked BERT (0.7)  & 65.3 & 38.0 & 50.9 & 43.5 & 58.9 & 24.2 & 32.5 & 27.7\\
  \hline
  \end{tabular}

    \centering
    \label{tab:ablation}
\end{table*}

\subsection{Effect of Hyper Parameter}

We present the results of Soft-Masked BERT on (the test data of) News Title to illustrate the effect of parameter and data size.

Table~\ref{tab:size} shows the results of Soft-Masked BERT as well as BERT-Finetune learned with different sizes of training data. One can find that the best result is obtained for Soft-Masked BERT when the size is 5 million, indicating that the more training data is utilized the higher performance can be achieved. One can also observe that Soft-Masked BERT is consistently superior to BERT-Finetune. 

A larger $\lambda$ value means a higher weight on error correction. Error detection is an easier task than error correction, because essentially the former is a binary classification problem while the latter is a multi-class classification problem. Table~\ref{tab:hyper} presents the results of Soft-Masked BERT in different values of hyper-parameter $\lambda$. The highest F1 score is obtained when $\lambda$ is 0.8. That means that a good compromise between detection and correction is reached.

\begin{table}[htb]
\caption{Impact of Different Values of $\lambda$} 
\label{tab:hyper} 
\includegraphics[width=8cm,height=3cm]{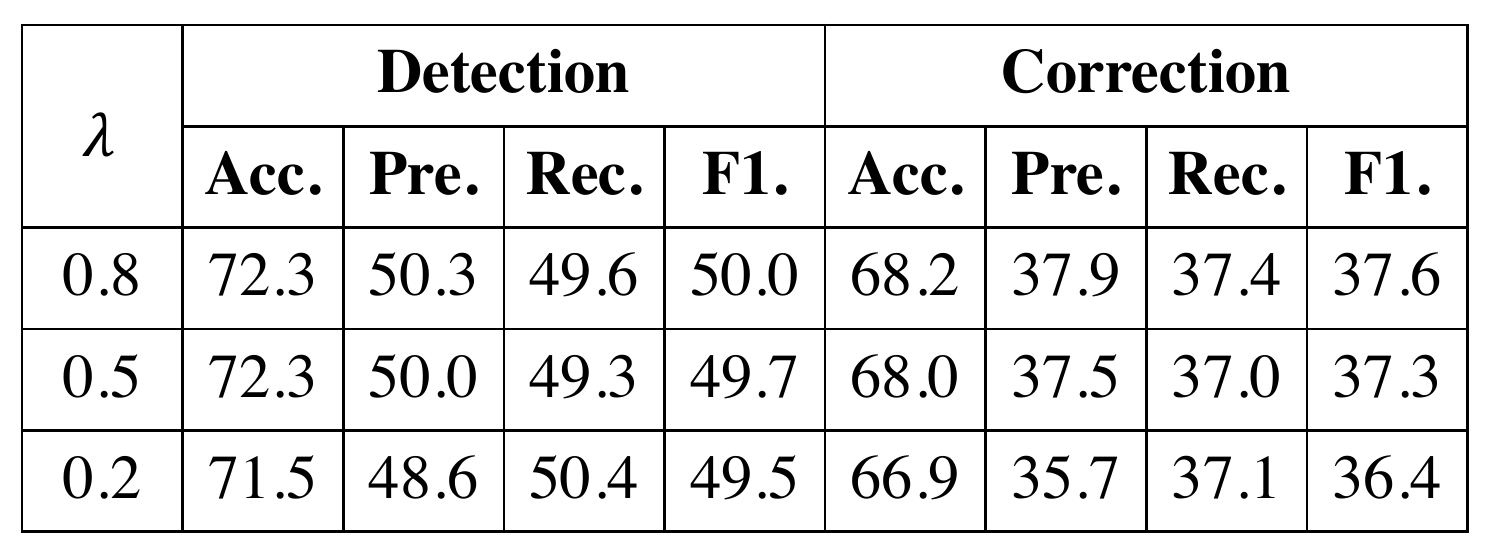}
\end{table}

\subsection{Ablation Study}
We carried out ablation study on Soft-Masked BERT on both datasets. Table~\ref{tab:ablation} shows the results on News Title. (We omit the results on SIGHAN due to space limitation, which have similar trends.) In Soft-Masked BERT-R, the residual connection in the model is removed. In Hard-Masked BERT, if the error probability given by the detection network exceeds a threshold (0.95, 0.9, 07), then the embedding of the current character is set to the embedding of the [MASK] token, otherwise the embedding remains unchanged. In Rand-Masked BERT, the error probability is randomized with a value between 0 and 1. We can see that all the major components of Soft-Masked BERT are necessary for achieving high performance. We also tried `BERT-Finetune + Force', whose performance can be viewed as an upper bound.  In the method, we let BERT-Finetune to only make prediction at the position where there is an error and  select a character from the rest of the candidate list. 
The result indicates that there is still large room for Soft-Masked BERT to make improvement.

\subsection{Discussions}

We observed that Soft-Masked BERT is able to make more effective use of global context information than BERT-Finetune. With soft masking the likely errors are identified, and as a result the model can better leverage the power of BERT to make sensible reasoning for error correction by referring to not only local context but also global context. For example, there is a typo in the sentence \begin{CJK*}{UTF8}{gbsn}`我会说一点儿，不过一个汉子也看不懂，所以我迷路了'\end{CJK*}(I can speak a little Chinese, but I don't understand man. So I got lost.). The word \begin{CJK*}{UTF8}{gbsn}`汉子'\end{CJK*}(man) is incorrect and should be written as \begin{CJK*}{UTF8}{gbsn}`汉字'\end{CJK*}(Chinese character). BERT-Finetune can not rectify the mistake, but Soft-Masked BERT can, because the error correction can only be accurately conducted with global context information.

We also found that there are two major types of errors in almost all methods including Soft-Masked BERT, which affect the performances. For statistics of errors, we sampled 100 errors from test set. We found that 67\% of errors require strong reasoning ability, 11\% of errors are due to lack of world knowledge, and the remaining 22\% of errors have no significant type. 

The first type of errors is due to lack of inference ability. Accurate correction of such typos requires stronger inference ability. For example, for the sentence \begin{CJK*}{UTF8}{gbsn}`他主动拉了姑娘的手, 心里很高心, 嘴上故作生气'\end{CJK*} (He intentionally took the girl's hand, and was very x, but was pretending to be angry.) where the incorrect word `x' is not comprehensible, there might be two possible corrections, changing \begin{CJK*}{UTF8}{gbsn}`高心'\end{CJK*} to \begin{CJK*}{UTF8}{gbsn}`寒心'\end{CJK*}(chilled) and changing \begin{CJK*}{UTF8}{gbsn}`高心'\end{CJK*} to \begin{CJK*}{UTF8}{gbsn} `高兴'\end{CJK*}(happy), while the latter is more reasonable for humans. One can see that in order to make more reliable corrections, the models must have stronger inference ability.

The second type of errors is due to lack of world knowledge. For example, in the sentence \begin{CJK*}{UTF8}{gbsn}`芜湖: 女子落入青戈江,众人齐救援'\end{CJK*} (Wuhu: the woman fell into the Qingge River, and people tried hard to rescue her.), \begin{CJK*}{UTF8}{gbsn}`青戈江'\end{CJK*} (Qingge River) is a typo of \begin{CJK*}{UTF8}{gbsn}`青弋江'\end{CJK*} (Qingyu River). Humans can 
discover the typo because the river in Wuhu city China is called Qingyu not Qingge. 
It is still very challenging for the existing models in general AI systems to detect and correct such kind of errors. 

\section{Related Work}

Various studies have been conducted on spelling error correction so far, which plays an important role in many applications, including search~\cite{DBLP:conf/coling/GaoLMQS10}, optical character recognition (OCR)~\cite{afli2016using}, and essay scoring~\cite{burstein1999automated}.

Chinese spelling error correction (CSC) is a special case, but is more challenging due to its conflation with Chinese word segmentation, which  received a considerable number of  investigations~\cite{yu2014overview,yu-li-2014-chinese,tseng2015introduction,wang2019confusionset}. Early work in CSC followed the  pipeline of error detection, candidate generation, and final candidate selection. Some researchers employed unsupervised methods using language models and rules~\cite{yu-li-2014-chinese,tseng2015introduction} and the others viewed it as a sequential labeling problem and employed conditional random fields or hidden Markov models~\cite{tseng2015introduction,zhang2015hanspeller++}. Recently, deep learning was applied to spelling error correction~\cite{guo2019spelling,wang2019confusionset}, and for example, a Seq2Seq model with BERT as encoder was employed~\cite{hong2019faspell}, which transforms the input sentence into a new sentence with spelling errors corrected.  


BERT~\cite{devlin2018bert} is a language representation model with Transformer encoder as its architecture. BERT is first pre-trained using a very large corpus in a self-supervised fashion (mask language modeling and next sentence prediction). Then, it is fine-tuned using a small amount of labeled data in a down-stream task. Since its inception BERT has demonstrated superior performances in almost all the language understanding tasks, such as those in the GLUE challenge~\cite{wang2018glue}. BERT has shown strong ability to acquire and utilize knowledge for language understanding. Recently, other language representation models have also been proposed, such as XLNET~\cite{yang2019xlnet}, Roberta~\cite{liu2019roberta}, and ALBERT~\cite{lan2019albert}. In this work, we extend BERT to Soft Masked BERT for spelling error correction and as far as we know no similar architecture was proposed before.

\section{Conclusion}
In this paper, we have proposed a novel neural network architecture for spelling error correction, more specifically Chinese spelling error correction (CSC). Our model called Soft-Masked BERT is composed of a detection network and a correction network based on BERT.  The detection network identifies likely incorrect characters in the given sentence and soft-masks the characters. The correction network takes the soft-masked characters as input and makes correction on the characters. The technique of soft-masking is general and potentially useful in other detection-correction tasks. Experimental results on two datasets show that Soft-Masked BERT significantly outperforms the state-of-art method of solely utilizing BERT. 
As future work, we plan to extend Soft-Masked BERT to other problems like grammatical error correction and explore other possibilities of implementing the detection network.

\bibliography{acl2020}

\begin{thebibliography}{17}
\expandafter\ifx\csname natexlab\endcsname\relax\def\natexlab#1{#1}\fi

\bibitem[{Afli et~al.(2016)Afli, Qiu, Way, and Sheridan}]{afli2016using}
Haithem Afli, Zhengwei Qiu, Andy Way, and P{\'a}raic Sheridan. 2016.
\newblock Using smt for ocr error correction of historical texts.
\newblock In \emph{Proceedings of the Tenth International Conference on
  Language Resources and Evaluation (LREC'16)}, pages 962--966.

\bibitem[{Burstein and Chodorow(1999)}]{burstein1999automated}
Jill Burstein and Martin Chodorow. 1999.
\newblock Automated essay scoring for nonnative english speakers.
\newblock In \emph{Proceedings of a Symposium on Computer Mediated Language
  Assessment and Evaluation in Natural Language Processing}, pages 68--75.
  Association for Computational Linguistics.

\bibitem[{Devlin et~al.(2018)Devlin, Chang, Lee, and
  Toutanova}]{devlin2018bert}
Jacob Devlin, Ming-Wei Chang, Kenton Lee, and Kristina Toutanova. 2018.
\newblock Bert: Pre-training of deep bidirectional transformers for language
  understanding.
\newblock \emph{arXiv preprint arXiv:1810.04805}.

\bibitem[{Gao et~al.(2010)Gao, Li, Micol, Quirk, and
  Sun}]{DBLP:conf/coling/GaoLMQS10}
Jianfeng Gao, Xiaolong Li, Daniel Micol, Chris Quirk, and Xu~Sun. 2010.
\newblock \href {https://www.aclweb.org/anthology/C10-1041/} {A large scale
  ranker-based system for search query spelling correction}.
\newblock In \emph{{COLING} 2010, 23rd International Conference on
  Computational Linguistics, Proceedings of the Conference, 23-27 August 2010,
  Beijing, China}, pages 358--366.

\bibitem[{Guo et~al.(2019)Guo, Sainath, and Weiss}]{guo2019spelling}
Jinxi Guo, Tara~N Sainath, and Ron~J Weiss. 2019.
\newblock A spelling correction model for end-to-end speech recognition.
\newblock In \emph{ICASSP 2019-2019 IEEE International Conference on Acoustics,
  Speech and Signal Processing (ICASSP)}, pages 5651--5655. IEEE.

\bibitem[{Hong et~al.(2019)Hong, Yu, He, Liu, and Liu}]{hong2019faspell}
Yuzhong Hong, Xianguo Yu, Neng He, Nan Liu, and Junhui Liu. 2019.
\newblock Faspell: A fast, adaptable, simple, powerful chinese spell checker
  based on dae-decoder paradigm.
\newblock In \emph{Proceedings of the 5th Workshop on Noisy User-generated Text
  (W-NUT 2019)}, pages 160--169.

\bibitem[{Lan et~al.(2019)Lan, Chen, Goodman, Gimpel, Sharma, and
  Soricut}]{lan2019albert}
Zhenzhong Lan, Mingda Chen, Sebastian Goodman, Kevin Gimpel, Piyush Sharma, and
  Radu Soricut. 2019.
\newblock Albert: A lite bert for self-supervised learning of language
  representations.
\newblock \emph{arXiv preprint arXiv:1909.11942}.

\bibitem[{Liu et~al.(2019)Liu, Ott, Goyal, Du, Joshi, Chen, Levy, Lewis,
  Zettlemoyer, and Stoyanov}]{liu2019roberta}
Yinhan Liu, Myle Ott, Naman Goyal, Jingfei Du, Mandar Joshi, Danqi Chen, Omer
  Levy, Mike Lewis, Luke Zettlemoyer, and Veselin Stoyanov. 2019.
\newblock Roberta: A robustly optimized bert pretraining approach.
\newblock \emph{arXiv preprint arXiv:1907.11692}.

\bibitem[{Martins and Silva(2004)}]{DBLP:conf/tal/MartinsS04}
Bruno Martins and M{\'{a}}rio~J. Silva. 2004.
\newblock \href {https://doi.org/10.1007/978-3-540-30228-5\_33} {Spelling
  correction for search engine queries}.
\newblock In \emph{Advances in Natural Language Processing, 4th International
  Conference, EsTAL 2004, Alicante, Spain, October 20-22, 2004, Proceedings},
  pages 372--383.

\bibitem[{Tseng et~al.(2015)Tseng, Lee, Chang, and
  Chen}]{tseng2015introduction}
Yuen-Hsien Tseng, Lung-Hao Lee, Li-Ping Chang, and Hsin-Hsi Chen. 2015.
\newblock Introduction to sighan 2015 bake-off for chinese spelling check.
\newblock In \emph{Proceedings of the Eighth SIGHAN Workshop on Chinese
  Language Processing}, pages 32--37.

\bibitem[{Wang et~al.(2018{\natexlab{a}})Wang, Singh, Michael, Hill, Levy, and
  Bowman}]{wang2018glue}
Alex Wang, Amanpreet Singh, Julian Michael, Felix Hill, Omer Levy, and Samuel~R
  Bowman. 2018{\natexlab{a}}.
\newblock Glue: A multi-task benchmark and analysis platform for natural
  language understanding.
\newblock \emph{arXiv preprint arXiv:1804.07461}.

\bibitem[{Wang et~al.(2018{\natexlab{b}})Wang, Song, Li, Han, and
  Zhang}]{wang2018hybrid}
Dingmin Wang, Yan Song, Jing Li, Jialong Han, and Haisong Zhang.
  2018{\natexlab{b}}.
\newblock A hybrid approach to automatic corpus generation for chinese spelling
  check.
\newblock In \emph{Proceedings of the 2018 Conference on Empirical Methods in
  Natural Language Processing}, pages 2517--2527.

\bibitem[{Wang et~al.(2019)Wang, Tay, and Zhong}]{wang2019confusionset}
Dingmin Wang, Yi~Tay, and Li~Zhong. 2019.
\newblock Confusionset-guided pointer networks for chinese spelling check.
\newblock In \emph{Proceedings of the 57th Annual Meeting of the Association
  for Computational Linguistics}, pages 5780--5785.

\bibitem[{Yang et~al.(2019)Yang, Dai, Yang, Carbonell, Salakhutdinov, and
  Le}]{yang2019xlnet}
Zhilin Yang, Zihang Dai, Yiming Yang, Jaime Carbonell, Ruslan Salakhutdinov,
  and Quoc~V Le. 2019.
\newblock Xlnet: Generalized autoregressive pretraining for language
  understanding.
\newblock \emph{arXiv preprint arXiv:1906.08237}.

\bibitem[{Yu and Li(2014)}]{yu-li-2014-chinese}
Junjie Yu and Zhenghua Li. 2014.
\newblock \href {https://doi.org/10.3115/v1/W14-6835} {{C}hinese spelling error
  detection and correction based on language model, pronunciation, and shape}.
\newblock In \emph{Proceedings of The Third {CIPS}-{SIGHAN} Joint Conference on
  {C}hinese Language Processing}, pages 220--223, Wuhan, China. Association for
  Computational Linguistics.

\bibitem[{Yu et~al.(2014)Yu, Lee, Tseng, and Chen}]{yu2014overview}
Liang-Chih Yu, Lung-Hao Lee, Yuen-Hsien Tseng, and Hsin-Hsi Chen. 2014.
\newblock Overview of sighan 2014 bake-off for chinese spelling check.
\newblock In \emph{Proceedings of The Third CIPS-SIGHAN Joint Conference on
  Chinese Language Processing}, pages 126--132.

\bibitem[{Zhang et~al.(2015)Zhang, Xiong, Hou, Zhang, and
  Cheng}]{zhang2015hanspeller++}
Shuiyuan Zhang, Jinhua Xiong, Jianpeng Hou, Qiao Zhang, and Xueqi Cheng. 2015.
\newblock Hanspeller++: A unified framework for chinese spelling correction.
\newblock In \emph{Proceedings of the Eighth SIGHAN Workshop on Chinese
  Language Processing}, pages 38--45.

\end{thebibliography}
\bibliographystyle{acl_natbib}

\end{document}